\documentclass[final]{cvpr}

\usepackage{times}
\usepackage{epsfig}
\usepackage{graphicx}
\usepackage{amsmath}
\usepackage{amssymb}

\usepackage{amsmath}
\usepackage{multirow}
\usepackage{amssymb}
\usepackage[ruled,linesnumbered]{algorithm2e}
\usepackage{bm}
\usepackage{booktabs}
\usepackage{float}
\usepackage{setspace}
\usepackage{lipsum}

\usepackage{hhline}
\usepackage{marvosym}
\usepackage{colortbl}

\usepackage[pagebackref=true,breaklinks=true,colorlinks,bookmarks=false]{hyperref}

\def\cvprPaperID{7211} %
\def\confYear{CVPR 2021}

\begin{document}

\title{Improving the Efficiency and Robustness of Deepfakes Detection through \\ Precise Geometric Features}

\author{Zekun Sun\textsuperscript{1} \quad Yujie Han\textsuperscript{1}\quad Zeyu Hua\textsuperscript{1}\quad \Letter Na Ruan\textsuperscript{1}\quad Weijia Jia\textsuperscript{1,2,3} \\

\small
\textsuperscript{1} Department of Computer Science and Engineering,
Shanghai Jiao Tong University, Shanghai, China \\
\small
\textsuperscript{2} Institute of Artificial Intelligence and Future Networks, Beijing Normal University (BNU Zhuhai), Guangdong, China
\\
\small
\textsuperscript{3} Key Lab of AI and Multi-Modal Data Processing,
BNU-HKBU United International College, Guangdong, China
\\
{\tt\small $\{$szk037, flora\_hua, jiawj$\}$@sjtu.edu.cn \quad borloch@outlook.com \quad naruan@cs.sjtu.edu.cn
}
}

\maketitle

\begin{abstract}
Deepfakes is a branch of malicious techniques that transplant a target face to the original one in videos, resulting in serious problems such as infringement of copyright,
confusion of information, or even public panic. Previous efforts for Deepfakes videos detection mainly focused on appearance features, which have a risk of being bypassed by sophisticated manipulation, also resulting in high model complexity and sensitiveness to noise. Besides, 
how to mine the temporal features of manipulated videos and exploit them is still an open question. 
We propose an efficient and robust framework named LRNet for detecting Deepfakes videos through temporal modeling on precise geometric features. A novel calibration module is devised to  enhance the precision of geometric features, making it more discriminative, and a two-stream Recurrent Neural Network (RNN) is constructed for sufficient exploitation of temporal features.
Compared to previous methods, our proposed method is lighter-weighted and easier to train. Moreover, our method has shown robustness in detecting highly compressed or noise corrupted videos.
Our model achieved 0.999 AUC on FaceForensics++ dataset. Meanwhile, it has a graceful decline in performance (-0.042 AUC) when faced with highly compressed videos.\footnote{
    Github: \url{https://github.com/frederickszk/LRNet}
}
\end{abstract}

\section{Introduction}
Due to the recent improvement of autoencoders and Generative Adversarial Networks (GAN) \cite{goodfellow2014generative}, synthetic videos are becoming unprecedentedly vivid and difficult to distinguish by either humans or machines. Deepfakes are the most flagrant models among those, which can change a person's identity in the video. Since face videos contain sensitive personal information, abuse of these techniques will grow into a menace. 
The advent of the forged speech of Barack Obama \cite{romano2018jordan} and manipulated pornographic videos of famous actresses \cite{spivak2019deepfakes} aroused great concern on the Internet.
Besides celebrities, ordinary people can also fall victim to Deepfakes 
on account of the abundant amount of video clips on social platforms and freely fetchable implementations of Deepfakes.
Therefore, how to detect Deepfakes videos becomes a matter of urgency.

Deepfakes detecting methods so far can be roughly classified into two types. The first type mainly focuses on defects in one single frame  \cite{mccloskey2018detecting, li2018detection, matern2019exploiting, li2019exposing, rossler2019faceforensics++, nguyen2019capsule, li2020xray}. 
And the second type takes temporal features into account \cite{ li2018ictu, agarwal2019protecting, sabir2019recurrent}. 
Some of the methods mentioned above mainly focus on non-essential defects of Deepfakes techniques (such as abnormal eye blinking or different colors of irises), which
in return stimulated the improvement of Deepfakes video synthesis.

In the context of an arms race between Deepfakes generation and detection techniques, 
there are several challenges need to be encountered. 
Firstly, advanced manipulation approaches urge detectors to uncover the intrinsic characteristics of Deepfakes videos, which cannot be easily disguised. 
Secondly, detectors should be more robust, enabling them to perform well on real-world detection missions. For instance, a lot of models \cite{chollet2017xception,afchar2018mesonet, li2020xray} witnessed severe performance drop on compressed videos, which reduces their effectiveness in application. Thirdly, the model simplicity should be taken into account. Current detection methods heavily rely on powerful Deep Convolutional Neural Networks (DCNNs) or data augmentation skills, which demands unendurable training costs. Also they are unfriendly for reproduction.

\begin{figure}[t]
  \centering
  \includegraphics[width=.99\columnwidth]{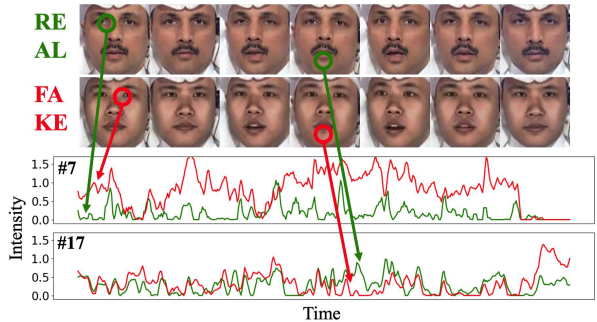}
  \caption{Action units (AU) intensity analysis for pristine and Deepfakes video sequence. AU indicate movements of individual facial muscles that make up the facial expression. We select the two most intense action units: \#7 (lid tightener) and \#17 (chin raiser). As we can see, although the fake sequence are too realistic to distinguish from appearance, we can still tell their differences on some subtle expressions, even though the faces in this two videos are performing the totally same action.}
  \label{fig:auAnalysis}
\end{figure}

We make a key observation that although manipulated face videos show high fidelity in a single frame, they still reveal some subtle but unnatural expressions or facial organs' movements. This is an inherent defect of Deepfakes techniques because forge videos are generated frame-by-frame, and no strong restriction is imposed on both individual behavior patterns and time continuity (as illustrated in Fig. \ref{fig:auAnalysis}).
To better capture these ``temporal artifacts”, also taking model robustness and simplicity into account, 
we opt for \emph{geometric features}, \eg , the shape and the position of facial organs. They can be more explicit for modeling facial dynamic behavior. Facial landmarks are a set of points outlining the contours of iconic facial parts, which are sufficient for describing geometric information and suitable for our framework.

Previous works have shown the potential of geometric features (especially the facial landmarks) in exposing synthetic face images or videos \cite{yang2019exposing, yang2019exposing2, agarwal2019protecting}. However, they utilized hand-crafted or complex correlation-based features for classification, which are not optimal for capturing all of distinguishable dynamic properties. In contrast, we design a two-stream Recurrent Neural Network to extract deep temporal features on landmarks sequence. Moreover, none of them have considered the influence brought by imprecise facial landmarks, which could be harmful to obtain more meaningful features. We devise a novel landmark calibration module to enhance the descriminative abilities of geometric features by reducing jittering, which enables us to combine the geometric and deep temporal features reliably and construct our detection framework dubbed Landmark Recurrent Network (LRNet).

Our framework LRNet achieves complementary strengths. On one hand, replacing face images with landmarks can be seen as an effective data dimensionality reduction. Compare to other deep-learning based model, it not only reduces model redundancy but also is more invariant to corruption in videos.
On the other hand, the deep RNNs help to enlarge the feature space and promote the expressiveness 
competence of facial landmarks. It strikes a better balance between cost and performance.

Our contributions can be concluded in three aspects:

\begin{itemize}
    \item We propose an efficient and robust framework to classify Deepfakes videos where we model temporal characteristics on precise geometric features.
    \item We introduce a novel plug-and-use landmark calibration module to promote the precision of geometric features and the effectiveness of temporal modeling while enabling our framework to be more flexible and reproducible. 
    \item We carry out extensive experiments to verify the efficiency and robustness of our method and also explore the influencing factors.
\end{itemize}

\section{Related Work}
\label{sec:relatedwork}
\subsection{Deepfakes Detection}
In this part, we introduce the current progress in the field of Deepfakes detection.

\textbf{Frame-level detection.}
Up to now, most of the efforts of the deepfakes detection are paid onto the single-frame based approach. 
Some of these techniques base on simple features selected manually. 
For instance, Matern et al. \cite{matern2019exploiting} focused on simple visual artifacts such as colors of irises, wired shadows on the face and missing details of eyes and teeth. 
The others turn to the deep features extract by DCNNs. Afchar et al. \cite{afchar2018mesonet} proposed
MesoNet
based on mesoscopic properties of images. R{\"o}ssler et al. \cite{rossler2019faceforensics++} successfully transferred Xception \cite{chollet2017xception} into deepfake detection task. Li et al. \cite{li2020xray} utilized an advanced architecture named HRNet
\cite{sun2019hrnet} to detect the blending boundary of Deepfakes manipulated images. These methods trade high costs for good performance thanks to powerful CNNs. Nonetheless, they are short of robustness or difficult to reproduce.

\textbf{Video-level detection.}
Recently, the intuition that videos contain more information than images greatly inspires the Deepfakes detection based on temporal features. Some geometric features-based schemes have made valuable attempts. Li et al.  \cite{li2018ictu} captured the abnormality of eye blinking frequency in fake videos. Yang et al. \cite{yang2019exposing} used outer landmarks and central landmarks to compose head direction and face direction respectively and detected the inconsistency between them. The manually selected features are less discriminative which limits their performance,
while we turn to exploit expressive deep temporal features. 
Appearance-based solutions are relatively more prevalent.
G{\"u}era et al. \cite{guera2018deepfake} proposed a framework utilizing a CNN to extract features from frames and a LSTM to process the features sequence. 
Sabir et al. \cite{sabir2019recurrent} adopted a similar architecture but replaced the LSTM with bidirectional Gated Recurrent Unit (GRU). These methods rely greatly on CNN as well, thus they suffer from similar problems as those frame-level detectors.

\subsection{Landmarks Detection}
Facial landmarks are representative and vital geometric features. It's detection methods have been widely studied for years. At the beginning, researchers proposed Active Appearance Model (AAM) \cite{cootes2001active} and Constrained Local Models (CLM) \cite{cristinacce2006clm}. Afterwards, the detection methods based on Cascaded Shape Regression (CSR) \cite{xiong2013csr1, kazemi2014one} achieved prominence. These methods make an initial estimate of the landmark locations then refine them iteratively through the ensemble of regression models (e.g., regression trees). They are adopted by widely-used open-source image processing 
repositories like \texttt{Dlib} \cite{king2009dlib}, which are easy to use and fast in detection speed.
Recently, abundant deep learning based models are devised such as Cascade CNN \cite{zhou2013cascadecnn}, Convolutional Pose Machine (CPM) \cite{wei2016cpm}, 
Convolutional Experts CLM \cite{zadeh2017ceclm}, 
etc. Some are also implemented by open-source toolkits like \texttt{Openface} \cite{baltrusaitis2018openface}. They have better performance while slower in speed.
Furthermore, sophisticated architectures are introduced to resolve the problems of face occlusion, extreme head pose, and so on. 

\begin{figure}[t]
  \centering
  \includegraphics[width=.75\columnwidth]{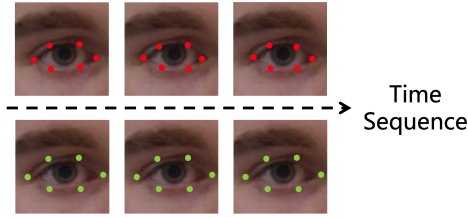}
  \caption{Comparison between accuracy and precision. Red landmarks (upper) are accurate but not precise. They jitter greatly even though they all attach to the contour. Green points (lower) are less accurate but precise, which describe dynamic properties better.}
  \label{fig:precision}
\end{figure}

It is essential to ensure the \emph{accuracy} and \emph{precision} of detected landmarks because they are the decisive features in geometric-based Deepfakes detection. Specifically, the term ``accurate” means the detection results have a low bias while “precise” refers to a low variance (as illustrated in Fig. \ref{fig:precision}).
The precision is relatively more important because the jittering noise would severely disturb the temporal modeling.
However, current landmarks detector mostly operate in frame-level and cannot achieve high precision. For this reason, we designed a calibration module to enhance the precision of landmarks detection results.

\section{Methodology}
\label{sec:method}
\begin{figure}[t]
  \centering
  \includegraphics[width=0.95\columnwidth]{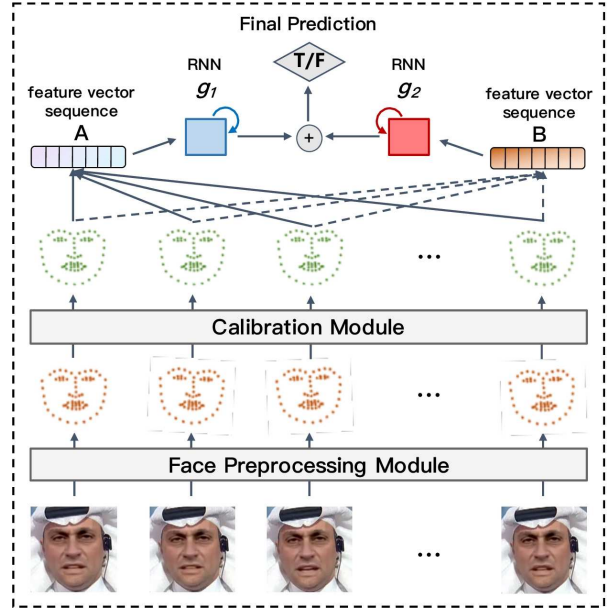}
  \caption{Overview of LRNet detection framework. the video to be detected is split into frames, and passed through the preprocessing procedure along with a carefully designed calibration module to obtain a sequence of more precise facial landmarks. 
  The subsequent embedding procedure embed landmarks into two types of feature vectors, and a two-stream RNN is used to mine the temporal information and judge its authenticity.   
}
  \label{fig:framework}
\end{figure}

Our proposed Deepfakes detection framework LRNet consists of four components (as shown in Fig. \ref{fig:framework}): face preprocessing module, calibration module, feature embedding procedure and RNN classifying procedure. It exposes manipulated faces by detecting abnormal facial movement patterns and time discontinuities.
Unlike most of the proposed methods that need to be trained end-to-end, our framework only requires the training of the RNNs part.
Details of our framework will be demonstrated below.

\subsection{Face Preprocessing}
The face preprocessing module extracts geometric information from face images. It consists of face detection, facial landmarks detection and landmarks alignment. 

In the beginning, face detection is performed on each frame of the video, and we retain the Region Of Interest (ROI) of the face. After cropping out the face images, we detect 68 facial landmarks on them, which outline the iconic profile on the face. Finally, we align landmark points to a preset position implemented by affine transformation.

Noted that our framework is flexible enough to decouple the preprocessing module (more specifically, the landmark detector). Firstly, the landmark detector can be implemented by pre-trained models without any additional training. Secondly, the performance of our whole framework does not heavily rely on the landmark detector. This property is guaranteed by our proposed calibration module which will be discussed below. And we demonstrate it in details by experiments in section \ref{sec:calib}.

\subsection{Landmark Calibration}
\begin{figure}[t]
  \centering
  \includegraphics[width=0.95\columnwidth]{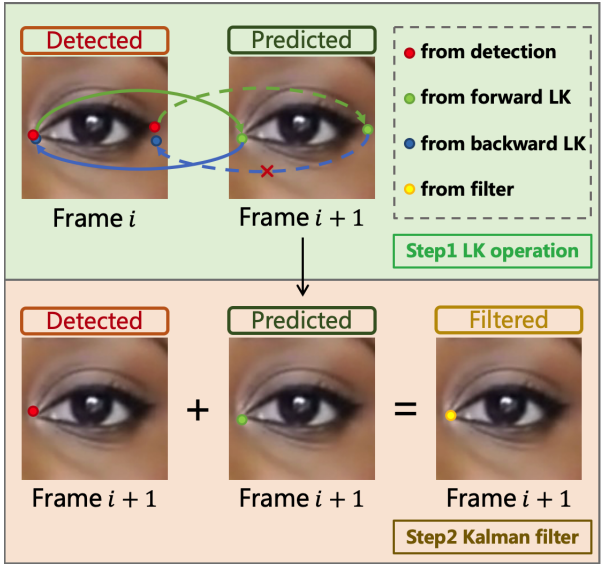}
  \caption{Detailed steps of the calibration module. The first step uses LK operation to track the landmarks point. Also a forward-backward check is performed to eliminate imprecise predictions. The second step uses Kalman filter to merge the detection and prediction results.
  }
  \label{fig:calibration}
\end{figure}

The landmarks extract through the preprocessing module can basically meet the demand for accuracy. However, they are far from precise for they are detected frame-by-frame. From our observation, the detected landmarks will have obvious jittering even if the face is almost not moving. Therefore, we proposed a novel calibration module 
to
tackle this problem (as illustrated in Fig. \ref{fig:calibration}). We use the successive frames to predict the latter positions of landmarks based on Lucas-Kanade optical flow calculation algorithm \cite{lucas1981iterative, baker2004lucas}. And valid predictions are merged with its corresponding detection results by a customized Kalman filter \cite{kalman1960new} to denoise and obtain the calibrated landmarks with higher precision. 

\subsubsection{Tracking}
For calibrating the landmarks, our intuition lies in that we can adjust their positions by matching small image patches around them. For this reason, Lucas-Kanade algorithm is suitable because it calculates the optical flow, the movement of several feature points between frames, essentially in the same way to this intuition. Motivated by works of Baker et al. \cite{baker2004lucas} and Dong et al. \cite{dong2018sbr}, we proposed a pyramidal Lucas-Kanade operation (dubbed LK operation below) to predict the landmark positions, in other words, track the landmarks. 

Given a small image patch $\mathbf{P}_i$ from frame$_i$ centered at $\mathbf{x}_{i} =[x, y]^\mathrm{T}$, where another same-size patch $\mathbf{P}_{i+1}$ from frame$_{i+1}$, we try to find a displacement vector $\mathbf{d} = [d_x, d_y]^\mathrm{T}$ to minimize the difference between $\mathbf{P}_i$ and $\mathbf{P}_{i+1}$, then we can obtain the tracking prediction $\mathbf{x}_{i+1} = \mathbf{x}_i + \mathbf{d}$. Therefore, we can calculate displacement vector $\mathbf{d}$ by minimizing
\begin{equation}
    \label{eq:initial}
    \sum_{\mathbf{x} \in \Omega} \alpha_{\mathbf{x}} \left\|\mathbf{P}_i(\mathbf{x} + \Delta \mathbf{d}) - \mathbf{P}_{i+1}(\mathbf{x} + \mathbf{d}) \right\|^2 ,
\end{equation}
where $\mathbf{d}$ is firstly initialized to be $[0, 0]^\mathrm{T}$. From eq. (\ref{eq:initial}) we can solve the $\Delta \mathbf{d}$ and iteratively update $\mathbf{d}$ by 
\begin{equation}
    \label{eq:update}
    \mathbf{d} \leftarrow \mathbf{d} + \Delta \mathbf{d}
\end{equation}
until convergence. In Eq. (\ref{eq:initial}), $\Omega$ refers to the set of all the positions in patch centered at $\mathbf{x}_{i}$, and $\alpha_{\mathbf{x}} = {\rm exp}(- \frac{\left\|\mathbf{x} - \mathbf{x}_{i} \right\|_2^2}{2\sigma^2})$, which is used to reduce the weights of locations far from the center and make a soft prediction.

According to the work of Baker et al. \cite{baker2004lucas}, we attain the final solution of Eq. (\ref{eq:initial}) that:
\begin{equation}
    \label{eq:solution}
    \Delta \mathbf{d} = \mathbf{H}^{-1} \sum_{\mathbf{x} \in \Omega} J(\mathbf{x}) \alpha_{\mathbf{x}}(\mathbf{P}_{i+1}(\mathbf{x} + \mathbf{d}) - \mathbf{P}_i(\mathbf{x} + 0)).
\end{equation}
In Eq. (\ref{eq:solution}), $\mathbf{H} = \mathbf{J}^\mathrm{T}\mathbf{A}\mathbf{J} \in \mathbb{R}^{2 \times 2}$ is the Hessian matrix. $\mathbf{J} \in \mathbb{R}^{C|\Omega| \times 2}$ is generated by vertically concatenating $J(\mathbf{x}) \in \mathbb{R}^{C \times 2}(\mathbf{x} \in \Omega) $, which is the Jacobian matrix of $\mathbf{P}_i(\mathbf{x} + 0)$. $C$ is the number of channels of $\mathbf{P}_i$ (3 for RGB images). $\mathbf{A}$ is a diagonal matrix whose elements are composed of $\alpha_{\mathbf{x}}$ to weight the corresponding Jacobian of $\mathbf{x}$ in $\mathbf{J}$. The advantage of this solution lies in that $\mathbf{P}_i(\mathbf{x})$ is constant during the iteration, thus complicated $\mathbf{J}$ and $\mathbf{H}^{-1}$ are only computed for once. 

Considering that LK operation is sensitive to the patch's size, 
We introduce pyramidal LK operation (detailedly described in Algorithm \ref{ag:LK}) that firstly downsamples the image several times (usually halve its size) to build pyramid representation of it, and perform simple LK operation on images of different size with the same patch size. 

\begin{algorithm}[t]
\caption{Pyramidal LK operation}
\label{ag:LK} 
\KwIn{former frame $\mathbf{F}_i$, latter frame $\mathbf{F}_{i+1}$, \quad\quad point in former frame to be tracked $\mathbf{x}_i$}
\KwOut{ predicted point in letter frame $\mathbf{x}_{i+1}$}
Build pyramid representation of $F_i$ and $F_{i+1}$: $\left\{ F_i^{L} \right\}_{L=0, ..., L_m}$, $\left\{ F_{i+1}^{L} \right\}_{L=0, ..., L_m}$\;
$\mathbf{g}^{L_m} \leftarrow [0, 0]^{\mathrm{T}}$\; 
\For{ $L=L_m; L\ge 0; L--$ }{
    $\mathbf{x}_{i}^{L} \leftarrow \mathbf{x}_{i}/2^L$\;
    Extract patch $P_i^L$ from $F_i^L$ centered at $\mathbf{x}_i^L$\;
    Compute the Jacobian matrix, $\mathbf{J}$ and $\mathbf{H}$ for $P_i^L$\;
    $\mathbf{d}^L \leftarrow [0, 0]^{\mathrm{T}}$\;
    \For{$i=1:max$}{
        Extract patch $P_{i+1}^L$ from $F_{i+1}^L$ centered at $\mathbf{x}_i^L + \mathbf{d}^L$\;
        Compute $\Delta \mathbf{d}^L$ by Eq. (\ref{eq:solution})\;
        Update $\mathbf{d}^L$ by Eq. (\ref{eq:update})\;
    }
    $\mathbf{g}^{L-1} \leftarrow 2(\mathbf{g}^L + \mathbf{d}^L)$\;
}
Obtain final prediction: $\mathbf{d} \leftarrow \mathbf{g}^0 + \mathbf{d}^0$\;
Return: $\mathbf{x}_{i+1} \leftarrow \mathbf{x}_{i} + \mathbf{d}$\;
\end{algorithm}

Noticed that LK operation is not always successful, thus a forward-backward check is introduced as shown in Fig. \ref{fig:calibration}. We perform a forward LK operation (green arrows and points) on the former frame, and a backward LK operation (blue arrows and points) on predicted points from the latter frame back to the former frame. The predicted point with a large difference between its original point and backward LK point will be discarded (dotted arrows).

\subsubsection{Denoising}
We discover from experimental results that LK operation can also bring in noise, which disturbs the stability of landmarks. Consequently, we devise a customized Kalman filter to integrate the information from both detections and predictions instead of only the LK operation results. 

Kalman filter estimates optimal landmark point $\mathbf{x}_{i+1}^{opt}$ in frame$_{i+1}$ through a weighted average of corresponding landmark detection result $\mathbf{x}_{i+1}^{det}$ and LK operation tracking prediction $\mathbf{x}_{i+1}^{pred}$. This procedure can be represented by:
\begin{equation}
    \label{eq:kalman}
    \mathbf{x}_{i+1}^{opt} = \mathbf{x}_{i+1}^{pred} + K_{i+1}\cdot(\mathbf{x}_{i+1}^{det} - \mathbf{x}_{i+1}^{pred} ),
\end{equation}
where $K_{i+1}$ is the Kalman gain when estimating $\mathbf{x}_{i+1}^{opt}$. It can be computed by:
\begin{equation}
    \label{eq:gain}
    K_{i+1} = \frac{P_{i+1}}{P_{i+1} + D_{i+1}},
\end{equation}
where $P_{i+1}$ is the variance (indicating the instability) of LK operation when predicting $\mathbf{x}_{i+1}^{pred}$, and $D_{i+1}$ is similarly the variance of landmark detection when detecting $\mathbf{x}_{i+1}^{det}$. 
Afterwards, we update next LK operation's variance $P_{i+2}$ by:
\begin{equation}
    \label{eq:updateP}
    P_{i+2} = (1 - K_{i+1})\cdot P_{i+1} + Q,
\end{equation}
where $Q$ is the inherent variance of LK operation.

However, it's difficult to calculate $P$ and $D$ because neither LK operation nor landmark detector is a simply-representable mathematical model. We thus propose the conception of \textbf{approximate relative variance} $D^{r}$ that:
\begin{equation}
    \label{eq:dr}
    D_{i+1}^r = \frac{\mathbf{x}_{i+1}^{det} - \mathbf{x}_i}{\mathbf{x}_{i+1}^{pred} - \mathbf{x}_{i}}.
\end{equation}
Due to the fact that each ground-truth landmark point in successive frame won't shift greatly, if the detection result have a apparent vibration, $D^{r}$ will bigger than 1. We then empirically set $Q=0.3$ according to visual effect in experiments, and replace $D$ with $D^{r}$ when computing Eq. (\ref{eq:gain}).

Our calibration module depends on the landmarks of frame$_1$ to calibrate the landmarks of frame$_2$. Then these optimized landmarks of frame$_2$ will be used to calibrate the frame$_3$ and so on. Given the lanmarks $\mathbf{L}_i = [\mathbf{x}_i^1, ..., \mathbf{x}_i^{68}]^\mathrm{T}$ and $\mathbf{L}_{i+1}$ extracted from frame $\mathbf{F}_i$ and $\mathbf{F}_{i+1}$ the overall procedure of landmark calibration 
is detailedly describe in Algorithm \ref{ag:calibration}. For simplicity we only express a single calibration step in it.

\begin{algorithm}[t]
\caption{Landmarks calibration}
\label{ag:calibration} 
\KwIn{$\mathbf{L}_i$, $\mathbf{L}_{i+1}$, $\mathbf{F}_i$, $\mathbf{F}_{i+1}$}
\KwOut{ Calibrated landmarks $\hat{\mathbf{L}_{i+1}}$}

\For{ $\mathbf{x}_i \in \mathbf{L}_i$}{
    $\mathbf{x}_{i+1} \leftarrow$ Algorithm \ref{ag:LK} ($\mathbf{F}_i$, $\mathbf{F}_{i+1}$, $\mathbf{x}_i$)\;
    $\tilde{\mathbf{x}_i} \leftarrow$ Algorithm \ref{ag:LK} ($\mathbf{F}_{i+1}$, $\mathbf{F}_{i}$, $\mathbf{x}_{i+1}$)\;
    Perform forward-backward check with $\mathbf{x}_i$ and $\tilde{\mathbf{x}_i}$\;
    \eIf(\tcc*[f]{Kalman filter}){ $\mathbf{x}_i$ \rm{is valid}}{
        Calculate $D_{i+1}^r$ by Eq. (\ref{eq:dr})\;
        Calculate $K_{i+1}$ by Eq. (\ref{eq:gain})\;
        Estimate $\mathbf{x}_{i+1}^{opt}$ by Eq. (\ref{eq:kalman})\;
        Update $P_{i+2}$ by Eq. (\ref{eq:updateP})\;
        $\hat{\mathbf{x}_{i+1}} \leftarrow \mathbf{x}_{i+1}^{opt}$\;
    }{
        $\hat{\mathbf{x}_{i+1}} \leftarrow$ corresponding $\mathbf{x}_{i+1} \in \mathbf{L}_{i+1}$ \;
    }
}
Return: $\hat{\mathbf{L}_{i+1}} = [\hat{\mathbf{x}_{i+1}^1}, ..., \hat{\mathbf{x}_{i+1}^{68}}]^\mathrm{T}$\;
\end{algorithm}

\subsection{Fake video classification}
The landmarks sequence extracted and calibrated in above steps are embedded into two types of feature vectors sequence, and then input to a two-stream RNN for fake video classification.

Each landmark point $\mathbf{x}^a$ can be represented by $\mathbf{x}^a = [x^a, y^a]^{\mathrm{T}}$, thus the first type of feature vector $\bm{\alpha}_i$ embedded from landmarks $\mathbf{L}_i = [\mathbf{x}_i^1, ..., \mathbf{x}_i^{68}]^\mathrm{T}$ can be generated by:
\begin{equation*}
    \bm{\alpha}_i = 
    \begin{bmatrix} x_i^1,& y_i^1,&x_i^2,&y_i^2,&... ,&x_i^{68},&y_i^{68} \end{bmatrix},
\end{equation*}
which can be seen as directly flatten from $\mathbf{L}_i$. Then the second type feature vector $\bm{\beta}_i$ can be computed by:
\begin{equation*}
\begin{aligned}
    \bm{\beta}_i 
    &=  \bm{\alpha}_{i+1} - \bm{\alpha}_{i} \\
    &=  \begin{bmatrix} x_{i+1}^1 - x_{i}^1,& ...,& y_{i+1}^{68} - y_i^{68} 
        \end{bmatrix},
\end{aligned}
\end{equation*}
representing the difference of landmarks' positions between successive frames.

Through embedding we obtain two feature vectors sequences $\mathbf{A} = [\bm{\alpha}_1, ..., \bm{\alpha}_n]^\mathrm{T}$ and $\mathbf{B} = [\bm{\beta}_1, ..., \bm{\beta}_{n-1}]^\mathrm{T}$. Afterwards, one RNN $g_1$ models facial shape movement pattern on $\mathbf{A}$, and the other RNN $g_2$ model landmarks difference pattern (or can be seen as the speed pattern, which is used to capture time discontinuity) on $\mathbf{B}$. Fully-connected layers are attached to each RNN's output to make its own prediction and the two streams are averaged as the final prediction. We conclude these prediction-making operation in one function $f(\cdot, \cdot)$. Therefore the final prediction, i.e., the real or fake possibility of a video clip, can be noted as:
\begin{equation}
    f(g_1(\mathbf{A}), g_2(\mathbf{B})).
\end{equation}
To perform the video-level detection, each video sample is segmented into clips with a fixed length. The predicted labels for clips are aggregated for the prediction of video.

\section{Experiments}
\label{sec:experiment}

In this section, we firstly declare the experiment settings. Then we evaluate the efficiency of our proposed LRNet framework on several benchmarks. Furthermore, we analyze the influencing factors of LRNet.
\subsection{Experiment Setting}
\subsubsection{Datasets}
Several challenging datasets have been proposed throughout the research progress of Deepfakes. To make the evaluation representative and comprehensive, we selected 4 typical datasets.

UADFV \cite{li2018ictu} contains 49 pristine videos and 49 manipulated videos. It represents the early dataset and adopted by a lot of classical works.

FaceForensic++ (FF++) \cite{rossler2019faceforensics++} contains 1000 videos as well as their manipulated version. Each video has a original version (raw), slightly-compressed version (c23) and heavily-compressed version (c40). It's the most typical recent dataset and has been widely adopted.

Celeb-DF \cite{li2020celeb} and DeeperForensics-1.0 (DF1.0)\cite{jiang2020deeperforensics} are two newly proposed datasets with high visual quality. Celeb-DF contains 5639 fake videos and 540 real videos, and DF1.0 contains 1000 real and corresponding fake videos similar to FF++. Each work also provide a benchmark which facilitates our evaluation.

\subsubsection{Parameters and implementation details}
In preprocessing step, we adopt \texttt{Dlib} \cite{king2009dlib} to carry out face and landmark detection (another detector \texttt{Openface}\cite{baltrusaitis2018openface} is adopt in the ablation study). In classification procedure, Each RNN in our two-stream network is bidirectional and consists of GRU (Gated Recurrent Unit) whose number of output units is set to be $k=64$. And two fully-connected layers with the number of units to be 64 and 2 are connected to RNN layer's output. A dropout layer with drop rate $dr_1 = 0.25$ is inserted between input and RNN, and another 3 dropout layers with $dr_2 = 0.5$ are used to separate the rest of the layers. 
These settings are partially based on existing reserach results \cite{sabir2019recurrent}.
In addition, we adopt an 8:2 dataset split, i.e., 80\% for training and 20\% for testing.
Each video is segmented into clips with a fixed length of 60, which sum up to 2 seconds when the fps is 30. 
We adopt Adam optimizer with $ lr = 0.001$, and batch size is set to be $1024$. This classification model will be trained up to 500 epochs.

\subsection{Performance Evaluation}
\subsubsection{General evaluation}

\begin{table}[t]
  \footnotesize
  \centering
  \setlength\tabcolsep{2.5pt}
    \begin{tabular}{c||ccc||cccc}
    \specialrule{0.1em}{0pt}{0pt}
    \multirow{2}[0]{*}{Methods} & \multicolumn{3}{c||}{Configurations} & \multicolumn{3}{c}{Testing Datasets} \\
     
    \hhline{~||---||---}      
            & Size  & Aug.  & Training & UADFV & FF++  & Celeb-DF\\
            
    \hline
    \hline
    Meso4 \cite{afchar2018mesonet} & 0.03 M & $\times$     & Unpub. & 84.3  & 84.7  & 54.8   \\
    FWA \cite{li2019exposing}  & 26 M   & $\surd$   & Unpub. & 97.4  & 80.1  & 56.9   \\
    DSP-FWA \cite{li2019exposing} & 28 M   & $\surd$     & Unpub. & 97.7  & 93.0  & \textbf{64.6}   \\
    Xception \cite{rossler2019faceforensics++}& 20.8 M & $\times$     & FF++  & 80.4  & 99.7  & 48.2   \\
    Capsule \cite{nguyen2019capsule}& 15 M   & $\times$     & FF++  & 61.3  & 96.6  & 57.5   \\
    CNN+RNN \cite{sabir2019recurrent} & 24.3 M & $\times$ & FF++ & 70.9 & 98.3 & 61.5 \\
    \hline
    \hline
    \textbf{LRNet (ours)} & 0.18 M & $\times$     & FF++  & \textbf{98.5} & \textbf{99.9} & 56.9 \\

    \specialrule{0.1em}{0pt}{10pt}
    \end{tabular}%
  \caption{General performance evaluation by AUC socres (\%) on different testing datasets. ``Aug." refers to if the method adopts data augmentation. Our proposed LRNet is relatively lightweight in the model's size and not in need of data augmentation, while performs the best on FF++.}
  \label{tab:general}
\end{table}

We firstly make a general evaluation of LRNet based on Celeb-DF benchmark \cite{li2020celeb}. Because a big part of currently proposed detection methods didn't open-source, making themselves difficult to be reproduced, we follow the evaluation setting of Celeb-DF benchmark that train our model on one dataset (mostly FF++) and test on different datasets. The evaluation metric is AUC scores (Area Under ROC Curve) and the results are shown in Table \ref{tab:general}. We only show the results of part of the best-performance methods. Our method achieves an almost full AUC score on its training dataset FF++ (99.9), showing that it can effectively capture the abnormal movements and discontinuities. Besides, it can also generalize to other datasets (unseen samples).

\subsubsection{Robustness to video compression}
We further test our methods robustness to video compression, which is overlooked by the majority of current works. On FF++, we compare the best detector on its benchmark, Xception, as well as newly-proposed and advanced X-Ray \cite{li2020xray}. Each detector is trained on original video (raw) and tested on three version of videos with different compression rates. On Celeb-DF, we used its benchmark settings that Xcpetion trained with FF++(c23) and FWA(DSP-FWA) trained with data augmentation. While our method directly train on FF++(raw).
The results are shown in Table \ref{tab:rob_comp}. We can draw from the results that the performance our methods is relatively more invariant to video compression.

\begin{table}[ht]
    \setlength\tabcolsep{8pt}
    \renewcommand\arraystretch{1.05}
    \small
    \centering
    \begin{tabular}{ccccc}
        \specialrule{0.1em}{0pt}{0pt}
        \multirow{2}[0]{*}{Methods} & \multicolumn{3}{c}{FF++} & \multirow{2}[0]{*}{Decline} \\
        \cline{2-4}          & raw   & c23   & c40   &  \\
        \hline
        Xception \cite{rossler2019faceforensics++} & 99.7  & 93.3  & 86.5  & 6.4/13.2 \\
        X-Ray \cite{li2020xray} & 99.1  & 87.3  & 61.6  & 11.8/37.5 \\
        \textbf{LRNet (ours)} & \textbf{99.9}  & \textbf{97.3}  & \textbf{95.7}  & \textbf{2.6}/\textbf{4.2} \\
        \hline
        \hline
        \multirow{2}[0]{*}{Methods} & \multicolumn{3}{c}{Celeb-DF} & \multirow{2}[0]{*}{Decline} \\
        \cline{2-4}         & raw   & c23   & c40   &  \\
        \hline
        Xception-c23 \cite{rossler2019faceforensics++} & \textbf{65.3}  & \textbf{65.5}  & 52.5  & \textbf{-0.2}/12.8 \\
        FWA \cite{li2019exposing}  & 56.9  & 54.6  & 52.2  & 2.3/4.7 \\
        DSP-FWA \cite{li2019exposing} & 64.6  & 57.7  & 47.2  & 6.9/17.4 \\
        \textbf{LRNet (ours)} & 56.9  & 56.3  & \textbf{55.4}  & 0.6/\textbf{1.5} \\
        \specialrule{0.1em}{0pt}{10pt}
    \end{tabular}%
    \caption{AUC scores (\%) of different methods when encountering video compression.
    }
    \label{tab:rob_comp}%
\end{table}%

\subsubsection{Robustness to video noise}

We in addition challenge our proposed LRNet on videos corrupted by different noise. We select DF1.0 benchmark, which is suitable for this evaluation. Because it provides this setting and tests on various advanced video-level detection methods that are ignored by other benchmarks. The results are shown in Table \ref{tab:noi_comp}. We can see that all of the methods perform well on the same training and testing dataset (include our LRNet who achieves 97.74\% acc. and 99.2\% AUC). While our methods suffer from the least performance decline when faced with noise.

\begin{table}[htb]
  \renewcommand\arraystretch{1.05}
  \setlength\tabcolsep{10pt}
  \small
  \centering
    \begin{tabular}{cccc}
        \specialrule{0.1em}{0pt}{0pt}
        \multirow{2}[0]{*}{Methods} & \multicolumn{2}{c}{Train/Test} & \multirow{2}[0]{*}{Decline} \\
        \cline{2-3}          & std/std & std/noise &  \\
        \hline
        \hline
        C3D \cite{tran2015lc3d}   & 98.50  & 87.63 & 10.87 \\
        TSN \cite{wang2016tsn}   & 99.25 & 91.50  & 7.75 \\
        I3D \cite{tran2015lc3d}   & \textbf{100.00} & 90.75 & 9.25 \\
        CNN+RNN \cite{sabir2019recurrent} & 100.00   & 90.63 & 9.37 \\
        Xception \cite{rossler2019faceforensics++} & 100.00   & 88.38 & 11.62 \\
        \hline
        \hline
        \textbf{LRNet (Ours)} & 97.74 & \textbf{96.83} & \textbf{0.91} \\
        \specialrule{0.1em}{0pt}{10pt}
        \end{tabular}
  \caption{Comparison of different method's robustness to video noise, which can are evaluated by binary classification accuracy (\%).
  ``std" refers to clean samples and ``noise" refers to samples with several types of strong noise as described in benchmark \cite{jiang2020deeperforensics}.}
  \label{tab:noi_comp}%
\end{table}%

\subsubsection{Efficiency in training}
To better demonstrate our framework's efficiency, we carry out evaluations over several representative baseline external methods and show the results in Table \ref{tab:cost}.
All models require similar pre-processing operations and our proposed LRNet consumes acceptable time, i.e., only 2h more than the basic requirement (6h). While our model is significantly faster and less memory intensive in training.

\begin{table}[t]
  \footnotesize
  \centering
  \setlength\tabcolsep{2.7pt}
    \begin{tabular}{c||cc||cccc}
    \hline
    \specialrule{0.1em}{0pt}{0pt}
    \multirow{2}[0]{*}{Model} & \multicolumn{2}{c||}{Pre-processing} & \multicolumn{4}{c}{Training} \\
\hhline{~||--||----} & Operations & Time  & \#Param & GPU   & DISK  & Time \\
    \hline
    \hline
    Xception\cite{chollet2017xception} & \cellcolor[rgb]{ .906,  .902,  .902} F+L+A & \cellcolor[rgb]{ .906,  .902,  .902} 6h & 20.8M & 12G   & 64G   & 21h \\
    X-Ray\cite{li2020xray} & \cellcolor[rgb]{ .906,  .902,  .902} F+L+A+D & \cellcolor[rgb]{ .906,  .902,  .902} 24h & 37.7M & $>$12G  & $>$180G & $>$30h \\
    CNN+RNN\cite{sabir2019recurrent} & \cellcolor[rgb]{ .906,  .902,  .902} F+L+A & \cellcolor[rgb]{ .906,  .902,  .902} 6h & 24.3M & 9G    & 64G   & 22.5h \\
    TSN\cite{wang2016tsn}   & \cellcolor[rgb]{ .906,  .902,  .902} F+L+A+O & \cellcolor[rgb]{ .906,  .902,  .902} 10h & 22.5M & $>$12G  & $>$120G & $>$30h \\
    \hline
    \hline
    \textbf{LRNet(Ours)} & \cellcolor[rgb]{ .906,  .902,  .902} \textbf{F+L+A+C} & \cellcolor[rgb]{ .906,  .902,  .902} \textbf{8h} & \textbf{0.18M} & \textbf{3G} & \textbf{1.1G} & \textbf{0.2h} \\
    \specialrule{0.1em}{0pt}{2pt}
    \end{tabular}%
  \caption{Quantitative comparisons for training cost (on FF++). The operations include face detection(F), landmark detection(L), alignment(A), data augmentation(D), optical flow(O) and proposed landmark calibration(C). \#Param refers to the trainable parameter size of the model. We also evaluate the GPU memory occupation (GPU), and disk memory occupied by training data (DISK).}
  \label{tab:cost}%
  \vspace{-0.5em}
\end{table}%

\subsection{Framework Analysis}

\subsubsection{Effect of calibration module}
\label{sec:calib}

\begin{figure}[tb]
  \centering
  \includegraphics[width=\columnwidth]{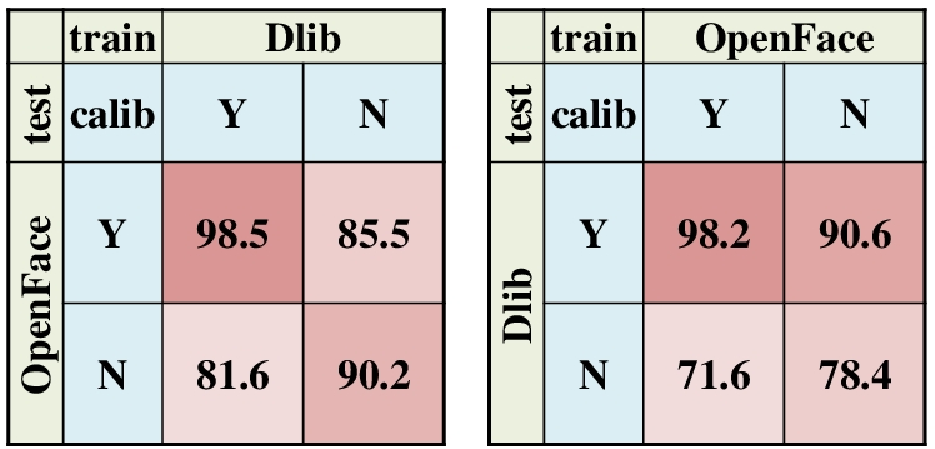}
  \caption{The confuse matrices for train on one kind of landmarks and test with the other. We evaluate the accuracy (\%) of each setting on FF++(raw). ``Y" means that detecting landmarks with our calibration module while ``N" means not.}
  \label{fig:confuse}
\end{figure}

\begin{figure}[tb]
  \centering
  \includegraphics[width=\columnwidth]{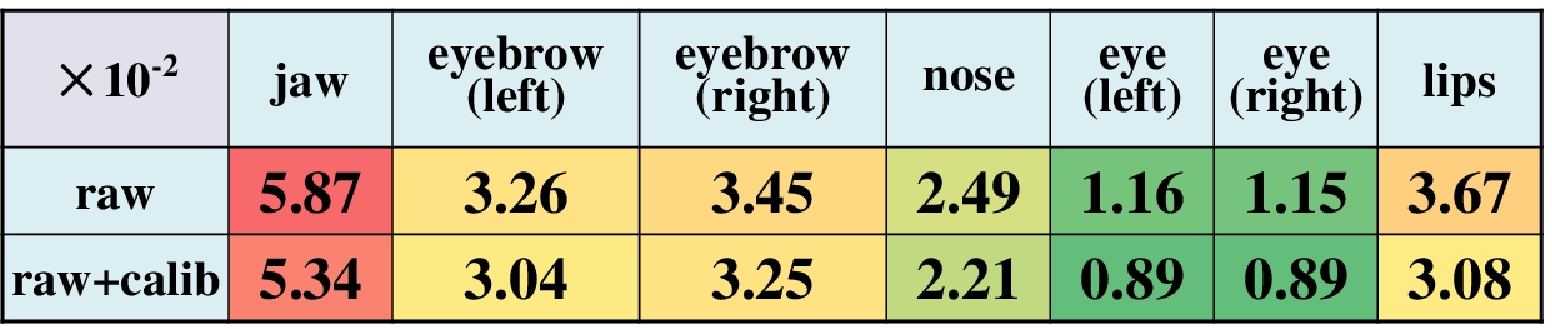}
  \caption{The average distance between the landmarks detected by different detectors (Dlib and OpenFace here). ``+calib" means utilize our calibration module. We merge the 68 landmarks into 7 groups by the organs they belong to.}
  \label{fig:landmarkDiff}
\end{figure}

Landmarks calibration is the core component of LRNet. We perform an ablation study on it (as shown in
Tabel \ref{tab:calib} 
). We can see that the calibration module enhances the overall performance of the detection framework as well as keeping its robustness.

We take a step further in evaluation. Firstly we try to detect landmarks with a better-performing CNN-based detector, \texttt{Openface} \cite{baltrusaitis2018openface}, and retrain the RNNs part of LRNet. The performances are very similar to the results of using \texttt{Dlib} and we do not list them in detail here. Then we explore the influence of not-retraining the model, i.e., feeding \texttt{Openface} landmarks into LRNet trained by \texttt{Dlib} landmarks, vice versa. As shown in Fig. \ref{fig:confuse}, only with the help of calibration module can it avoid a performance drop. The reason lies in that model trained by calibrated landmarks can better capture the abnormal facial movements instead of the noise brought by landmark detectors. We further demonstrate this by calculating the differences between landmarks from different detectors as shown in Fig. \ref{fig:landmarkDiff}. Calibration module helps shorten the gap of different landmarks detections, bring them closer to the ground-truth positions.

\begin{table}[tb]
    \setlength\tabcolsep{8pt}
    \renewcommand\arraystretch{1.05}
    \small
    \centering
    \begin{tabular}{c||cc||cc}
        \specialrule{0.1em}{0pt}{0pt}
        \multirow{2}[0]{*}{Methods} & \multicolumn{2}{c||}{FF++(raw)} & \multicolumn{2}{c}{FF++(c40)} \\
        \hhline{~||--||--} 
        & Acc   & AUC   & Acc  & AUC \\
        \hline
        \hline
        \cellcolor[rgb]{ .906,  .902,  .902} LRNet  &\cellcolor[rgb]{ .906,  .902,  .902}99.7  &
        \cellcolor[rgb]{ .906,  .902,  .902}99.9  &\cellcolor[rgb]{ .906,  .902,  .902}91.2  & 
        \cellcolor[rgb]{ .906,  .902,  .902}95.7 \\
        \hhline{-||--||--} 
        w/o Kalman filter & 98.5  & 99.4  & 87.2  & 94.3 \\
        w/o calibration & 92.8  & 97.4  & 84.3  & 92.6 \\
        \specialrule{0.1em}{0pt}{10pt}
    \end{tabular}%
  \caption{Ablation study of calibration module in LRNet by evaluating the binary detection accuracy (\%) and AUC scores (\%). All the models are only trained on FF++(raw). 
  }
  \label{tab:calib}%
\end{table}%

\subsubsection{Effect of network architecture}
We demonstrate the effectiveness of our two-stream RNN architecture by comparing it with only using its one stream (as shown in Table \ref{tab:arch}). The results show that this structure has a superior ability. The information from two roads promotes each other, where time discontinuity clues detected by $g_2$ contribute more to final accuracy and abnormal movements 
messages recognized by $g_1$ provide more robustness to the prediction.

\begin{table}[t]
    \setlength\tabcolsep{8pt}
    \renewcommand\arraystretch{1.05}
    \small
    \centering
    \begin{tabular}{c||cc||cc}
        \specialrule{0.1em}{0pt}{0pt}
        \multirow{2}[0]{*}{Methods} & \multicolumn{2}{c||}{FF++(raw)} & \multicolumn{2}{c}{FF++(c40)} \\
        \hhline{~||--||--}          
        & Acc   & AUC   & Acc   & AUC \\
        \hline
        \hline
        \cellcolor[rgb]{ .906,  .902,  .902}LRNet ($g_1+g_2$) &\cellcolor[rgb]{ .906,  .902,  .902}99.7  &
        \cellcolor[rgb]{ .906,  .902,  .902}99.9  & \cellcolor[rgb]{ .906,  .902,  .902}91.2    & 
        \cellcolor[rgb]{ .906,  .902,  .902}95.7 \\
        \hhline{-||--||--} 
        $g_1$    & 83.4  & 89.3  & 80.4  & 86.1 \\
        $g_2$    & 98.3  & 99.2  & 85.2  & 93.9 \\
        \specialrule{0.1em}{0pt}{10pt}
    \end{tabular}%
    \caption{Ablation study of network architecture by evaluating the binary detection accuracy (\%) and AUC scores (\%). $g_1$ refers to the RNN which models abnormal facial movements and $g_2$ refers to the other RNN which captures time discontinuity.}
    \label{tab:arch}%
\end{table}%

\subsubsection{Influence of input length}

\begin{figure}[tb]
  \centering
  \includegraphics[width=0.85\columnwidth]{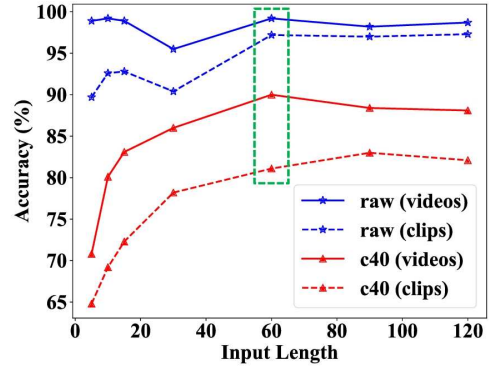}
  \caption{Accuracy (\%) of LRNet when the input length varies. All the models are trained on FF++(raw). 
  ``(clips)" refers to the sample-wise detection accuracy and ``(videos)" is the result of video-level classification.
    }
  \label{fig:inputLength}
\end{figure}

Input length refers to how many successive frames we feed it into the model as a single sample. We evaluate different input lengths with other conditions controlled (shown in Fig. \ref{fig:inputLength}). Despite the fact that input length hardly affects the performance when training and testing on the same dataset, a suitable input length (60 adopted by us) can improve both the effectiveness and robustness on 
different data distribution (e.g., compressed samples).

\section{Discussion}
\label{sec:discussion}
In this section, we discuss the our limitations and future works at first. From general evaluation results, there is still room for improvement in the generalizability of our framework. Besides, it's difficult to interpret the temporal features captured by our model and visualize the difference of movement pattern between real and fake faces. In future we will carry out a more in-depth research and analysis of these abnormal dynamic patterns of different face manipulation techniques, and then promote the model's generalizability.

We also expound the effect of appearance and geometric features. From the results of current works, appearance features are high-dimensional, more expressive to generalize, but less robust and high-cost. While geometric features are more robust, low-cost, but harder to generalize. So it's a trade-off of performance (especially generalization ability) and cost. While we make great efforts to promote the efficiency of Deepfakes detection only rely on the geometric features, it's worthy to explore if we can combine the appearance and geometric together at the same time avoiding high cost to improve the efficiency.

\section{Conclusion}
\label{sec:conclusion}
Deepfakes are huge threats to human society and their rapid development is calling for efficient solutions. Our work reveals that integration of facial landmarks and temporal features can be a fast and robust test of Deepfakes. We also explore how to enhance the landmark detection results and make full use of temporal features. We found that the facial geometric information and its dynamic characteristic are essential clues and worth exploring in future work for a more efficient and robust in-wild Deepfakes detection.

{\small
\bibliographystyle{ieee_fullname}
\bibliography{cvpr}
}

\end{document}